\documentclass[twocolumn]{article} 

\usepackage{multicol}  
\usepackage[letterpaper, top=0.6in, bottom=0.5in, inner=0.75in, outer=0.75in, headsep=0.3in, footskip=0.3in]{geometry}

\usepackage{titlesec}
\titleformat{\section}{\large\bfseries}{\thesection}{1em}{}
\titleformat{\subsection}{\normalsize\bfseries}{\thesubsection}{1em}{}

\usepackage{graphicx}         
\usepackage{amsmath, amssymb} 
\usepackage{hyperref}          
\usepackage{natbib}          
\usepackage{geometry}         
\usepackage{listings}
\usepackage{pgfplots}
\usepackage{amssymb} 
\usepackage{pifont}  
\usepgfplotslibrary{fillbetween}
\usepackage{algorithm}
\usepackage{algorithmic}
\usepackage{amsmath} 
\usepackage{appendix}
\usepackage{booktabs} 
\usepackage{caption}  

\title{\textbf{VoxRep: Enhancing 3D Spatial Understanding in 2D Vision-Language Models via Voxel Representation}}

\author{
    Alan Dao (Gia Tuan Dao)\textsuperscript{1}, Norapat Buppodom\textsuperscript{1} \\
    Menlo Research \\
    \texttt{alan@menlo.ai, norapat@menlo.ai} \\
    \textsuperscript{1}Equal contribution.
}

\date{\today}  

\begin{document}

\maketitle
\begin{abstract}
Comprehending 3D environments is vital for intelligent systems in domains like robotics and autonomous navigation. Voxel grids offer a structured representation of 3D space, but extracting high-level semantic meaning remains challenging. This paper proposes a novel approach utilizing a Vision-Language Model (VLM) to extract "voxel semantics"—object identity, color, and location—from voxel data. Critically, instead of employing complex 3D networks, our method processes the voxel space by systematically slicing it along a primary axis (e.g., the Z-axis, analogous to CT scan slices). These 2D slices are then formatted and sequentially fed into the image encoder of a standard VLM. The model learns to aggregate information across slices and correlate spatial patterns with semantic concepts provided by the language component. This slice-based strategy aims to leverage the power of pre-trained 2D VLMs for efficient 3D semantic understanding directly from voxel representations.
\noindent 
\end{abstract}

\section{Introduction}
\label{sec:introduction}

The ability to interpret and reason about the three-dimensional world is fundamental for advanced AI applications. Voxel representation, which discretizes 3D space into a regular grid, provides a structured format suitable for computational analysis \cite{zhou2017voxelnetendtoendlearningpoint}. While effective for representing occupancy and volume, deriving rich semantic understanding—identifying what objects are present, their attributes like color, and their precise locations—directly from the raw voxel data requires sophisticated interpretation.

Existing approaches often involve complex 3D convolutional networks operating directly on the voxel grid or methods focused on processing unstructured point clouds \cite{chen2023leveraginglargevisuallanguage}. We introduce a different paradigm for extracting "voxel semantics." Inspired by slice-based volumetric data analysis like CT scanning \cite{WikipediaVoxel}, we propose processing the 3D voxel grid by decomposing it into a sequence of 2D slices along a chosen axis (e.g., Z-axis). These 2D slices, capturing the cross-sectional information at different depths, are then treated as image inputs for a standard, pre-trained Vision-Language Model (VLM) \cite{bai2025qwen25vltechnicalreport, yao2024minicpmvgpt4vlevelmllm}.

\begin{figure}
    \centering
    \includegraphics[width=1\linewidth]{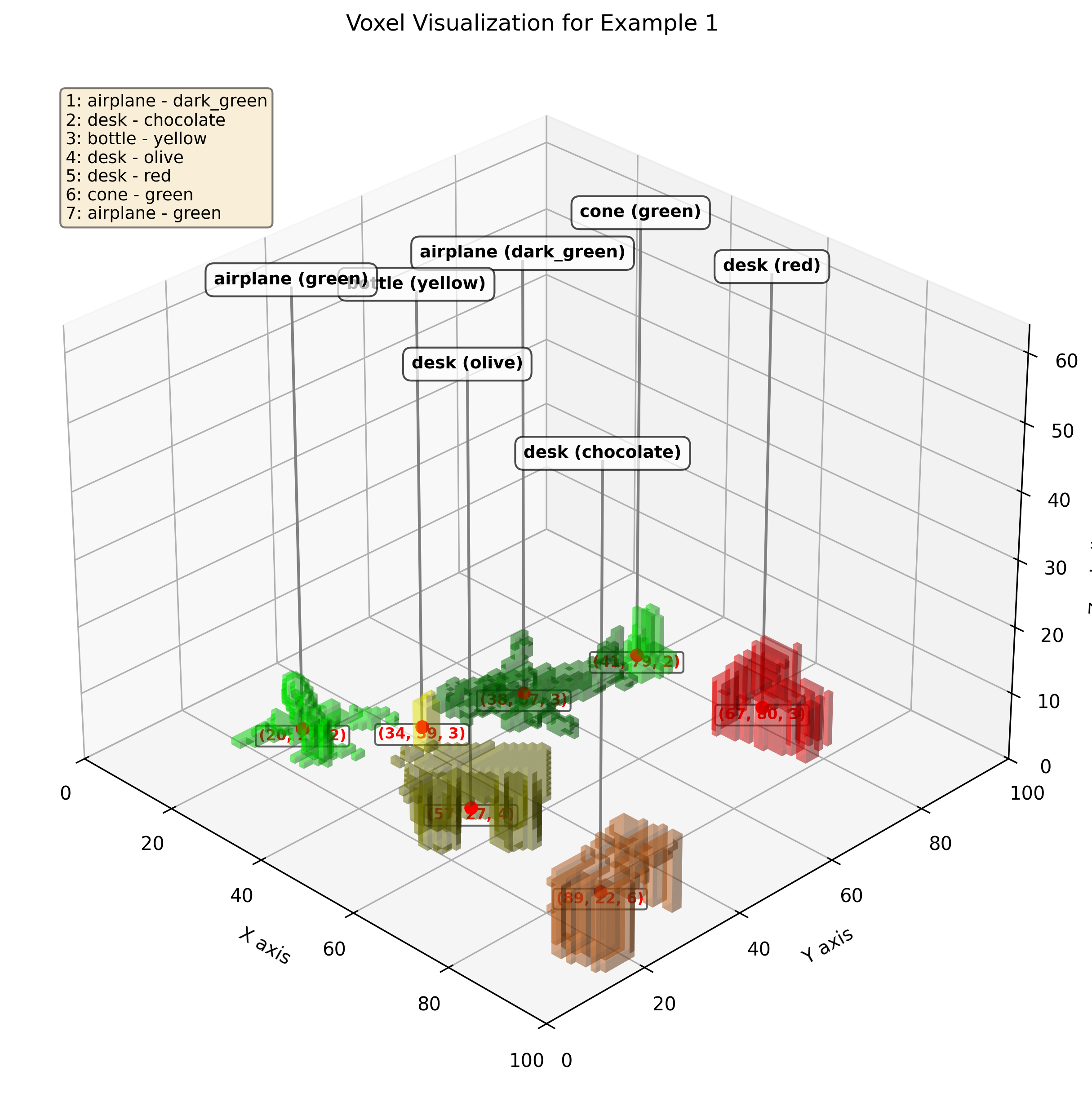}
    \caption{Put black cube onto green cube}
    \label{fig:concept-demo}
\end{figure}

The VLM's vision encoder processes each slice, and mechanisms are employed to aggregate features across the sequence of slices, allowing the model to reconstruct a contextual understanding of the 3D structure. The language component of the VLM guides the interpretation, mapping the aggregated visual features to semantic labels for objects, colors, and locations within the original voxel volume. We hypothesize that this slice-based approach can effectively leverage the powerful capabilities of existing 2D VLMs, which are often pre-trained on vast datasets, to achieve efficient and accurate voxel semantic extraction, thereby enhancing 3D scene understanding \cite{zhi2025lscenellmenhancinglarge3d,chandhok2024scenegptlanguagemodel3d,fu2024scenellmextendinglanguagemodel}.

We review the relevant background on voxel representation and scene understanding, detail the proposed slice-based VLM architecture, discuss the expected outcomes, advantages, and limitations of this specific approach, and outline future research avenues.

\begin{itemize}
    \item We demonstrate that general-purpose 2D Vision-Language Models (VLMs) can learn voxel representations through a data adaptation strategy that flattens 3D voxel grids into 2D image slices.
    \item We show that voxel representation learning can be effectively achieved at the vision-encoder level, enabling structured semantic extraction from voxel-based environments.
    \item Our approach leverages pre-trained VLMs without requiring expensive 3D-specific training, offering a scalable and efficient method for voxel-based scene understanding.
\end{itemize}
\section{Related Work}
\label{sec:related_work}

\subsection{Fundamentals of Voxel Representation}

Voxel representation involves discretizing 3D space into a regular grid structure \cite{wang20243drepresentationmethodssurvey}, where each volumetric element, or voxel, can store attributes like occupancy, color, or density. The conversion of geometric data into this grid format is known as voxelization. This approach is advantageous for representing volumetric detail and simplifies grid-based computations. However, a significant drawback is the high memory consumption associated with finer resolutions [1, 6], as storage needs scale cubically with grid dimensions \cite{arbore2024hybridvoxelformatsefficient}. Despite these trade-offs, the structured data facilitates various manipulation techniques. Notably, slice-based editing treats the volume as a stack of 2D images for modification \cite{WikipediaVoxel}, offering a precedent for our input processing methodology.

\subsection{3D Scene Understanding}

Scene understanding seeks to interpret 3D data beyond geometry, identifying objects, relationships, and context \cite{zhi2025lscenellmenhancinglarge3d,chandhok2024scenegptlanguagemodel3d}. It involves levels from detection to recognition \cite{zhi2025lscenellmenhancinglarge3d} and full understanding \cite{zhi2025lscenellmenhancinglarge3d}. Rich 3D data, like voxels, is crucial for this detailed comprehension \cite{luo2024differentiablevoxelizationmeshmorphing}, enabling understanding of functionality and affordances \cite{thengane2025foundationalmodels3dpoint}.

\subsection{Vision-Language Models and 3D Data Adaptation}

VLMs connect visual inputs with language. While some research explores native 3D VLMs, a common strategy involves adapting 3D data for use with powerful, readily available \cite{daxberger2025mmspatialexploring3dspatial}, pre-trained 2D VLMs \cite{chen2024spatialvlmendowingvisionlanguagemodels}. Point clouds might be projected into 2D views, or, as we propose, volumetric data can be sliced. Our approach falls into this category, aiming to reformat 3D voxel information into a sequence of 2D inputs compatible with standard VLM image encoders.

\section{Methodology}
\label{sec:methodology}

Our approach leverages the capabilities of the \textbf{Gemma 3 Vision-Language Model (VLM)} \cite{gemma3report}, an Encoder-Decoder model. A key factor in selecting Gemma 3 is its vision encoder's ability to process high-resolution image inputs, specifically up to \textbf{896 $\times$ 896 pixels} \cite{gemma3report}. This corresponds to approximately \textbf{800,000 pixels} ($896^2 \approx 802,816$). Our input 3D voxel grid has dimensions $100 \times 100 \times 16$, representing a total of \textbf{160,000 voxels}. The large input capacity of the Gemma 3 encoder significantly exceeds the number of voxels in our input space. This substantial pixel budget is crucial, providing ample room to represent the information from all voxel slices within a single 2D image, even after padding and resizing each slice (to $224 \times 224$) and tiling them together. Therefore, a core aspect of our methodology is a preprocessing pipeline designed to transform the input 3D voxel grid into this specific $896 \times 896$ 2D format, making it compatible with the Gemma 3 encoder. Following the encoding of this processed visual input, the Gemma 3 language decoder generates the structured semantic representation, termed "voxel semantics", from the encoded features. The subsequent sections detail this preprocessing pipeline and the structure of the semantic output.
\begin{figure}
    \centering
    \includegraphics[width=1\linewidth]{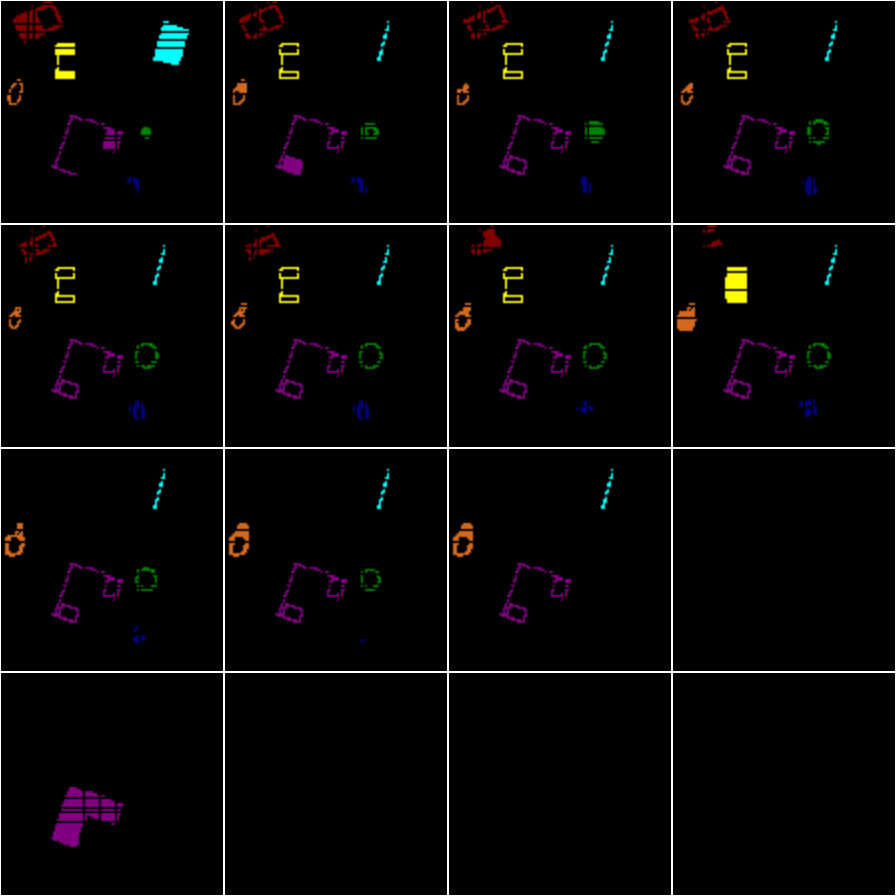}
    \caption{A sample of a sliced 3d voxels}
    \label{fig:voxel_slices}
\end{figure}
\subsection{Input Voxel Representation}
\label{subsec:input_voxel}

The input to our system is a discretized 3D scene represented as a voxel grid. Specifically, we consider a voxel space with dimensions $W \times H \times D$, where $W=100$, $H=100$, and $D=16$. Each voxel $(x, y, z)$ within this grid can store relevant properties of the space it occupies. For this work, we assume voxels primarily encode occupancy information and potentially color attributes (e.g., RGB values), particularly for occupied voxels. We operate under the assumption that relevant objects are primarily situated on the ground plane (i.e., at low $z$ values), although the processing handles all slices.

\subsection{Voxel Grid Preprocessing for VLM Encoder}
\label{subsec:preprocessing}

To leverage powerful pre-trained 2D vision encoders commonly found in VLMs (such as those used in models like Gemini), we transform the 3D voxel grid into a 2D image format through a series of steps:

\begin{enumerate}
    \item \textbf{Slicing:} The $100 \times 100 \times 16$ voxel grid is sliced along the depth (Z) axis. This decomposes the 3D volume into $D=16$ individual 2D slices, each with dimensions $100 \times 100$. Each slice represents a cross-section of the voxel space at a specific depth $z$.

    \item \textbf{Padding and Resizing:} Each $100 \times 100$ slice is padded to $112 \times 112$ pixels. It is then resized to $224$ pixels using standard image bilinear interpolation so that each voxel occupies four pixels. This specific dimension ($224$) is chosen based on common VLM architectural components where input images are often tiled or patched with $14 \times 14$ embedding layer

    (e.g., an $896 \times 896$ image might be processed as $4 \times 4$ patches of $224 \times 224$).
    

    \item \textbf{Tiling:} The $D=16$ padded slices, each of size $224 \times 224$, need to be arranged into a single 2D image grid that meets the target input dimensions of the VLM encoder, specified as $896 \times 896$. Since $896 / 224 = 8$, the target input requires an $4 \times 4$ grid of $224 \times 224$ patches. The slices are tiled in a predefined order (e.g., row-major order based on their original depth $z$). This results in a final composite 2D image of size $896 \times 896$ pixels. The channel dimension of this image corresponds to the information stored in the voxels (e.g., 3 channels for RGB color).
\end{enumerate}

This processed $896 \times 896$ image serves as the input to the vision encoder component of our VLM.

\subsection{Encoder-Decoder VLM Architecture}
\label{subsec:vlm_architecture}

We employ a standard Encoder-Decoder architecture:
\begin{itemize}
    \item \textbf{Vision Encoder:} Takes the preprocessed $896 \times 896$ image as input. It processes this tiled representation of the voxel space to generate a sequence of embedding vectors capturing the visual features. We leverage a pre-trained vision encoder compatible with this input size.
    \item \textbf{Decoder:} A language model (or a sequence-to-sequence model) takes the encoded visual features as input (potentially along with task-specific prompts or queries) and generates the structured semantic output.
\end{itemize}

\subsection{Output: Voxel Semantics Extraction}
\label{subsec:output_semantics}

While the Gemma 3 vision encoder processes the tiled 2D representation, it learns rich features that implicitly contain semantic information about the original 3D voxel space. The subsequent Gemma 3 language decoder is then tasked with \textbf{extracting and structuring} this learned information into a predefined format, which we term the "voxel semantics" output.

This output is designed as a structured representation, specifically a JSON list, where each element corresponds to a detected object instance within the original $100 \times 100 \times 16$ voxel grid. Each object entry contains the following key-value pairs:

\begin{itemize}
    \item \texttt{"id"}: A unique identifier assigned to the detected object instance (e.g., \texttt{"0"}, \texttt{"1"}, ...).
    \item \texttt{"color"}: A textual description of the object's dominant color (e.g., \texttt{"dark\_green"}, \texttt{"blue"}, \texttt{"grey"}). This is derived from the color information embedded within the voxels corresponding to the object.
    \item \texttt{"description"}: The semantic shape of the object (e.g., \texttt{"bowl"}, \texttt{"table"}, \texttt{"cube"}) drawn from the ModelNet40 dataset~\cite{wu20153d}, as detailed in Appendix~\ref{app:selected_categories}
    \item \texttt{"number\_of\_occupied\_voxel"}: An integer representing the approximate volume or density of the object, measured by the count of voxels it occupies within the original grid.
    \item \texttt{"voxel\_coords\_center"}: A nested JSON object specifying the approximate geometric center of the object within the original $100 \times 100 \times 16$ coordinate system. It contains:
        \begin{itemize}
            \item \texttt{"x"}: The center coordinate along the X-axis.
            \item \texttt{"y"}: The center coordinate along the Y-axis.
            \item \texttt{"z"}: The center coordinate along the Z-axis (depth).
        \end{itemize}
\end{itemize}

An example of the output format for a single detected object is:
\begin{lstlisting}
[
  {
    "id": "0",
    "color": "dark_green",
    "description": "bowl",
    "number_of_occupied_voxel": 1539,
    "voxel_coords_center": {"x": 71, "y": 64, "z": 5}
  }
  % ... potentially more objects in the list
]
\end{lstlisting}

This structured JSON output makes the semantic information learned by the model readily accessible and interpretable for downstream tasks requiring object identification, localization, and attribute recognition within the voxel space. The decoder essentially acts as a structured reporter for the semantic insights gained by the encoder.

\section{Experiments}
\label{sec:experiments}

\subsection{Experimental Setup}
\label{subsec:setup}

We trained the \textbf{Gemma 3 Vision-Language Model} \cite{gemma3report} to evaluate its capability for extracting voxel semantics. The training dataset consisted of \textbf{20,000 synthetic data points}. Each data point was generated by creating a $100 \times 100 \times 16$ voxel grid containing objects randomly sampled, positioned, oriented, and filled in color from the \textbf{ModelNet40 dataset} \cite{wu20153d}. Objects were primarily placed on the ground plane (low z-values) within the voxel space.

As detailed in the Methodology (Section~\ref{sec:methodology}), each $100 \times 100 \times 16$ voxel grid, containing occupancy and color information, was preprocessed into an $896 \times 896$ tiled 2D image representation suitable for the Gemma 3 vision encoder. The model was trained using standard optimization techniques (details omitted here for brevity) to predict the structured JSON output described in Section~\ref{subsec:output_semantics}, capturing the ID, description, color, voxel count, and center coordinates for each object identified within the voxel grid.

\subsection{Evaluation Metrics}
\label{subsec:metrics}

To assess the model's learning progress over training steps, we monitored the following metrics, calculated by comparing the model's predicted JSON output against the ground-truth annotations for each example in a validation set:

\begin{itemize}
    \item \textbf{Avg Center Distance}: The average Euclidean distance (in voxel units) between the predicted `voxel\_coords\_center` and the ground-truth center for matched object pairs. Lower is better.
    \item \textbf{Color Accuracy}: The proportion of matched objects where the predicted `color` string exactly matches the ground-truth color label. Higher is better.
    \item \textbf{Desc Accuracy} (Description Accuracy): The proportion of matched objects where the predicted `description` string (object category) exactly matches the ground-truth ModelNet40 label. Higher is better.
    \item \textbf{Avg Voxel Count Diff}: The average absolute difference between the predicted `number\_of\_occupied\_voxel` and the ground-truth voxel count for matched object pairs. Lower is better.
    \item \textbf{Avg Mismatch Per Example}: The Mean Absolute Error (MAE) between the \textit{number} of objects predicted by the model for a given example and the actual \textit{number} of ground-truth objects present in that example's scene. Calculated as $\frac{1}{N} \sum_{i=1}^{N} |\text{predicted\_count}_i - \text{true\_count}_i|$, where $N$ is the number of examples in the validation set. Lower values indicate the model is more accurately predicting the quantity of objects present.
\end{itemize}

We determine matched object pairs by finding the closest predicted and ground-truth objects based on Euclidean distance. Each matched object pair is formed by pairing a predicted object with the nearest ground-truth object. We ensures each predicted and ground-truth object is matched at most once.

\subsection{Results and Analysis}
\label{subsec:results}

We evaluated the model checkpoints at regular intervals from 200 to 1100 training steps. The evolution of the performance metrics is visualized using line charts in Figure~\ref{app:line_charts}. The raw numerical results corresponding to these charts are provided in Appendix~\ref{app:results_table}.

Analyzing the trends shown in Figure~\ref{app:line_charts} (and detailed in Appendix~\ref{app:results_table}), we observe the following:

\begin{itemize}
    \item \textbf{Localization Improvement (Avg Center Distance):} The model demonstrates significant improvement in localizing objects. The \textbf{Avg Center Distance} starts high (26.05 voxels at 200 steps) and decreases consistently throughout training, indicating increasingly accurate predictions of object centers. The most rapid improvement occurs between 400 and 600 steps, after which the rate of improvement slows. By 1100 steps, the average distance drops to 9.17 voxels, showing substantial learning of spatial information.

    \item \textbf{Attribute Recognition (Color \& Voxel Count):}
        \begin{itemize}
            \item \textbf{Color Accuracy} shows strong learning. Starting at only 0.22, it rises sharply to 0.73 by 600 steps and continues to improve gradually, reaching 0.78 by 1100 steps. This suggests the model effectively learns to map voxel color patterns to the correct descriptive labels.
            \item The \textbf{Avg Voxel Count Diff} mirrors the localization improvement. It starts very high (438.46) and decreases dramatically, particularly between 400 and 600 steps, eventually stabilizing around 182-187 towards the end. This indicates the model becomes progressively better at estimating the spatial extent or volume of the detected objects as it learns their geometry.
        \end{itemize}

    \item \textbf{Object Classification (Desc Accuracy):} Object description accuracy also improves steadily but lags behind color accuracy. Starting at 0.18, it reaches 0.50 by 500 steps and plateaus around 0.55-0.58 from 1000 steps onwards. While demonstrating learning, achieving high accuracy on object classification appears more challenging than color or location tasks within this experimental setup, potentially due to the visual similarity between some ModelNet40 categories or the limitations imposed by the voxel representation and 2D projection.

    \item \textbf{Detection Quality (Avg Mismatch Per Example):} This metric displays a more complex trend. It initially decreases substantially from 2.81 (200 steps) to a minimum of 0.59 (600 steps), indicating the model quickly learns to reduce spurious detections or gross localization errors. However, after 700 steps, the mismatch rate slightly increases again, hovering between 0.7 and 1.03. This later increase might suggest that as the model becomes more sensitive overall, it might start producing more borderline or slightly inaccurate detections in complex scenes, or struggle with resolving ambiguities, even as other metrics stabilize or improve slowly.
\end{itemize}

In summary, the experimental results clearly demonstrate the feasibility of training the Gemma 3 VLM, using our proposed voxel-to-2D preprocessing, to extract meaningful semantic information directly from voxel grids. The model shows strong learning curves for object localization, color recognition, and volume estimation, particularly within the first 600-700 training steps. Object classification accuracy improves but remains a key challenge. The performance generally begins to plateau after approximately 1000 steps, suggesting reasonable convergence within the observed training window. The slight increase in mismatches during later stages warrants further investigation into potential overfitting or difficulties with complex object arrangements.

\section{Discussion}
\label{sec:discussion}

The experimental results presented in Section~\ref{sec:experiments} 
demonstrate the viability of our proposed VoxRep approach for extracting semantic information from 3D voxel grids using a standard 2D Vision-Language Model (VLM). The core hypothesis—that a VLM can learn to interpret a tiled sequence of 2D voxel slices as a representation of a 3D scene—is largely supported by the findings.

\subsection{Interpretation of Results}
The model showed significant and relatively rapid improvement in localizing objects (\textit{Avg Center Distance}) and recognizing basic attributes like color (\textit{Color Accuracy}) and volume (\textit{Avg Voxel Count Diff}). This suggests that the Gemma 3 vision encoder, despite being pre-trained on natural 2D images, can effectively aggregate spatial features across the tiled slices to build an implicit 3D understanding sufficient for these tasks. The strong performance in color accuracy, reaching 0.78, indicates the model readily maps voxel color information within the slices to semantic labels. The substantial decrease in \textit{Avg Voxel Count Diff} implies the model learns to associate the spatial extent of activated voxels across slices with object volume.

Object classification (\textit{Desc Accuracy}), however, proved more challenging, plateauing at a moderate accuracy (0.58). This suggests that while the slice-based representation preserves enough information for localization and basic attributes, recognizing fine-grained geometric details necessary for distinguishing between similar object categories (like those in ModelNet40) might be hindered. The projection onto 2D slices potentially loses crucial 3D shape cues, or the VLM's inherent 2D biases make complex 3D shape interpretation difficult from this format.

The trend observed in \textit{Avg Mismatch Per Example} is noteworthy. The initial sharp decrease indicates the model quickly learns to identify distinct object presences and suppress spurious detections. However, the slight increase after 700 steps, while other metrics stabilize or improve, might suggest limitations in handling ambiguity or segmentation in more complex configurations as the model becomes more sensitive overall. It could point towards difficulties in separating closely spaced objects or correctly interpreting partial object views that span multiple slices.

\subsection{Strengths and Implications}
The primary strength of VoxRep lies in its ability to leverage powerful, readily available, pre-trained 2D VLMs for a 3D understanding task without requiring specialized 3D network architectures or extensive 3D pre-training. This offers a pathway towards more efficient and scalable 3D semantic extraction, benefiting from the continuous advancements in 2D vision and language modeling. The approach is conceptually simple and repurposes existing tools for a new domain.

\subsection{Limitations}
Several limitations should be acknowledged:
\begin{enumerate}
    \item \textbf{Information Loss:} The process of slicing the 3D volume into 2D images inherently discards some explicit information about connectivity along the slicing axis (Z-axis in this case). While the model learns to infer these connections, direct 3D convolutions might capture such relationships more naturally.
    \item \textbf{Representation Sensitivity:} The chosen voxel resolution (100x100x16) and the specific slicing axis impact the information available in the 2D representation. Performance might vary with different resolutions or slicing strategies.
    \item \textbf{Data Domain:} The experiments were conducted on synthetic, clean ModelNet40 data. Performance on real-world, noisy, and potentially sparse 3D data (e.g., from LiDAR or depth sensors) remains untested and might pose significant challenges.
    \item \textbf{Scene Complexity:} The dataset primarily featured isolated objects. The approach's effectiveness in complex, cluttered scenes with multiple interacting objects and occlusions needs further investigation. The tiling strategy might struggle to represent very deep or complex spatial layouts within the fixed 2D grid.
    \item \textbf{Fixed Input Size:} The method relies on fitting the tiled slices into the VLM's fixed input resolution (896x896). This imposes constraints on the original voxel grid dimensions (W, H, D) and the per-slice resolution after padding.
\end{enumerate}

\subsection{Future Work}
Future research should address these limitations. Testing VoxRep on real-world 3D datasets (e.g., ScanNet, KITTI) is crucial. Exploring alternative slicing strategies (e.g., multi-axis slicing) or more sophisticated tiling/aggregation mechanisms within the VLM could improve information capture. Investigating the impact of different VLM architectures and scaling the approach to handle larger, higher-resolution voxel grids are also important directions. Additionally, incorporating explicit mechanisms to handle scene clutter and inter-object relationships could enhance performance on more realistic tasks.
\section{Conclusion}
\label{sec:conclusion}

This paper introduced VoxRep, a novel method for enabling 2D Vision-Language Models to perform 3D semantic understanding directly from voxel representations. By systematically slicing a 3D voxel grid along one axis, padding and tiling these slices into a single 2D image compatible with standard VLM encoders, we demonstrated that a pre-trained model like Gemma 3 can learn to extract meaningful "voxel semantics"—including object identity, color, location, and volume.

Our experiments showed promising results, particularly in object localization, color recognition, and volume estimation, validating the core concept that 2D VLMs can infer 3D spatial and semantic information from this structured 2D representation. While object classification proved more challenging, the overall success highlights the potential of leveraging the power of large-scale pre-trained 2D models for 3D tasks, potentially offering a more efficient alternative to developing specialized 3D architectures.

Despite limitations related to potential information loss during slicing and the reliance on synthetic data, VoxRep presents a viable and scalable approach for bridging the gap between 2D foundation models and 3D voxel-based scene understanding. Future work will focus on evaluating the method on real-world data, exploring enhancements to the representation and aggregation strategy, and scaling the approach to handle more complex scenes and higher resolutions, further paving the way for effective VLM deployment in 3D environments.

\bibliographystyle{plainnat} 
\bibliography{bibliography} 


\clearpage
\appendix
\onecolumn
\appendix
\section{Results}

\begin{table}[h]
    \centering
    \begin{tabular}{|c|c|c|c|c|c|}
        \hline
        Steps & Avg Center Distance & Color Accuracy & Desc Accuracy & Avg Voxel Count Diff & Avg Mismatch Per Example \\
        \hline
        200  & 26.0534 & 0.22 & 0.18 & 438.46 & 2.81 \\
        300  & 24.2406 & 0.25 & 0.22 & 439.78 & 1.01 \\
        400  & 21.8985 & 0.38 & 0.32 & 369.32 & 0.66 \\
        500  & 15.8066 & 0.62 & 0.50 & 256.19 & 0.65 \\
        600  & 11.5417 & 0.73 & 0.56 & 206.91 & 0.59 \\
        700  & 10.3202 & 0.73 & 0.56 & 203.53 & 0.70 \\
        800  & 11.3598 & 0.72 & 0.53 & 201.23 & 0.82 \\
        900  & 10.1206 & 0.75 & 0.54 & 192.82 & 1.03 \\
        1000 & 9.4486  & 0.77 & 0.55 & 187.25 & 0.73 \\
        1100 & 9.1669  & 0.78 & 0.58 & 182.55 & 0.82 \\
        \hline
    \end{tabular}
    \caption{Training Results Over Steps}
    \label{app:results_table}
\end{table}

\begin{figure}
    \centering
    \includegraphics[width=1\linewidth]{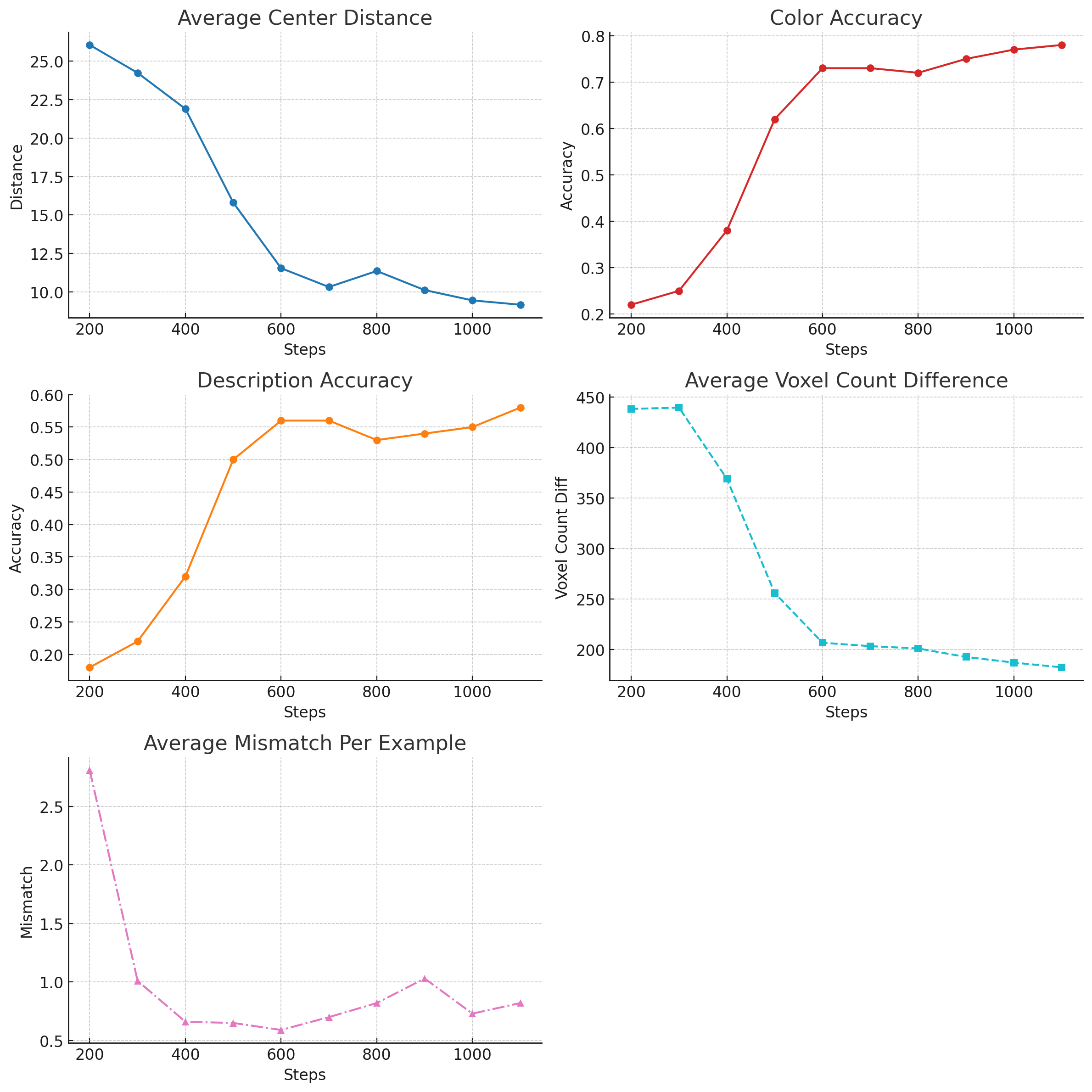}
    \caption{Performance Metrics Evolution during Train-ing. Line charts showing Avg Center Distance, ColorAccuracy, Description Accuracy, Avg Voxel Count Dif-ference, and Avg Mismatch Per Example plotted against training steps (200 to 1100}
    \label{app:line_charts}
\end{figure}

\section{Selected Categories}\label{app:selected_categories}

We selected the following object categories from the ModelNet40 dataset~\cite{wu20153d} for our synthesis data  pipeline:

\begin{itemize}
\item toilet
\item airplane
\item bathtub
\item bottle
\item bowl
\item cone
\item cup
\item desk
\item guitar
\item laptop
\item plant
\item sofa
\item stool
\item tent
\item toilet
\end{itemize}

\end{document}